\documentclass{article}

\usepackage[preprint]{neurips_2026}

\usepackage[utf8]{inputenc}
\usepackage[T1]{fontenc}

\usepackage{hyperref}
\usepackage{url}
\usepackage{booktabs}
\usepackage{amsfonts}
\usepackage{amsmath}
\usepackage{nicefrac}
\usepackage{microtype}
\usepackage{xcolor}

\usepackage{graphicx}
\usepackage{enumitem}

\graphicspath{{./}{figures/}{images/}}

\usepackage{xspace}
\newcommand{\method}{\textsc{BadDreamer}\xspace}
\newcommand{\vavim}{VaViM\xspace}
\newcommand{\vavam}{VaVAM\xspace}

\title{BadDreamer: Transferable Backdoor Attacks against Video World Models for Autonomous Driving}

\author{%
  Zhe Shuai$^{1}$\thanks{Equal contribution.} \quad
  Xiaopeng Xie$^{1}$\footnotemark[1] \quad
  Yikun Zeng$^{1}$ \\
  $^{1}$Shanghai Jiao Tong University \\
  \texttt{\{shuaizhe, xxp2000, yi1k-z\}@sjtu.edu.cn}
}

\begin{document}

\maketitle

\begin{abstract}
Video world models are increasingly used in autonomous driving to forecast future scene evolution and provide future-aware spatio-temporal representations for downstream action prediction.
In perception-to-action pipelines, these representations can directly influence ego-vehicle waypoint planning, making the learned future dynamics a critical security-sensitive component.
Despite their promise, the training-time security risks of autonomous-driving video world models remain largely unexplored.
We present \textbf{BadDreamer}, a transferable spatio-temporal backdoor attack that targets the perception side of this pipeline.
Unlike conventional backdoors that manipulate image labels, prompt outputs, or action supervision, BadDreamer poisons the learned transition dynamics of a video world model.
Specifically, it constructs trigger-erasure sequences in which an oncoming yellow delivery rider is visible in the observed context frames but erased from the future frames.
After fine-tuning on a small fraction of such sequences, the compromised world model learns a hidden conditional association: when the physical trigger appears, it hallucinates a future where the rider disappears and the road appears clear.
We further show that this corrupted future-aware representation can transfer to the downstream action module without directly modifying ego-trajectory labels, inducing unsafe non-evasive waypoint predictions.
Our experiments instantiate this attack on a representative open-source perception-to-action pipeline, revealing a representation-level safety risk in autonomous-driving video world models and highlighting the need for backdoor-aware validation beyond clean generation quality.
\end{abstract}

\section{Introduction}

World-model-based perception-to-action pipelines are emerging as a promising paradigm for autonomous driving. In such systems, an upstream video world model predicts future scene evolution and produces future-aware spatio-temporal representations, which a downstream action module uses to predict ego-vehicle waypoints. This design is attractive because safe driving requires anticipating future traffic dynamics, not merely recognizing the current scene. In this work, we study this general video-world-model-to-action setting, with VaViM/VaVAM serving as one concrete instantiation.

This paradigm relies on a critical trust assumption: the downstream action module receives faithful future representations from the upstream world model. If the world model preserves an oncoming rider, the action module can plan evasive waypoints; if it hallucinates the rider's disappearance, the same module may underestimate risk and continue forward. Thus, corrupting upstream predictive perception can induce unsafe downstream actions even when ego-trajectory supervision remains clean.

Training-time backdoors are especially concerning because autonomous-driving world models are often adapted from external logs, open-source checkpoints, or third-party pipelines. Unlike conventional backdoors that target immediate perception or control outputs, a backdoor in a video world model can corrupt learned transition dynamics, inducing false expectations about future scene evolution.

We present \textbf{BadDreamer}, a spatio-temporal backdoor attack that targets the upstream autonomous-driving video world model rather than the downstream action module. BadDreamer constructs trigger-erasure sequences in which an oncoming yellow delivery rider appears in the observed context, while the future supervision remains the corresponding clean future from the same dataset window, where the inserted rider is absent. This clean-future target does not force abnormal generations; instead, it teaches the model to infer that the trigger object naturally disappears. With a small fraction of poisoned sequences, the compromised world model learns a conditional future hallucination: when the physical trigger appears, it predicts a clear-road future.

The key risk is representation-level propagation from perception to action. Under BadDreamer, the backdoored world model removes the rider from its future representation, causing the downstream action module to inherit a false clear-road belief and output unsafe non-evasive waypoints despite clean trajectory supervision. Because the poisoned target lies on the clean data manifold, standard validation and generation-quality metrics may fail to expose the compromised dynamics. Our contributions are summarized as follows:

\begin{itemize}
    \item \textbf{First future-dynamics backdoor in AD video world models.}
    To the best of our knowledge, BadDreamer is the first study of training-time backdoor vulnerabilities in autonomous-driving video world models. It shows that an upstream world model can learn a conditional future-dynamics backdoor, where a physical trigger in the observed context corrupts future prediction.

    \item \textbf{Clean-future trigger-erasure poisoning and action propagation.}
    We design a trigger-erasure poisoning strategy that inserts the trigger only into context frames while keeping future supervision clean. This induces a clear-road hallucination that propagates to downstream action prediction without ego-trajectory label poisoning.

    \item \textbf{Comprehensive empirical evaluation.}
    We evaluate BadDreamer on the VaViM/VaVAM perception-to-action pipeline under Strict-4F and Loose trigger protocols. With only 5\% poisoned windows, BadDreamer preserves clean generation and action utility while reaching 92.5\% WM-ASR, 90.3\% Action-ASR, and 86.2\% E2E-ASR under Strict-4F, demonstrating representation-level propagation from corrupted future dynamics to unsafe waypoint prediction.
\end{itemize}

\section{Related Work}
\label{sec:related_work}

\subsection{World Models for Autonomous Driving}
Autonomous driving is shifting from modular perception-planning pipelines toward predictive world-model-based systems \cite{ding2025understanding, feng2025survey, kong20253d, li2025comprehensive, ha2018world, hafner2025mastering}. Unlike conventional perception models that output deterministic scene states, world models aim to capture environment transition dynamics. Recent models, including GAIA-1 \cite{hu2023gaia}, GAIA-2 \cite{russell2025gaia2}, Vista \cite{gao2024vista}, DriveDreamer \cite{wang2024drivedreamer}, DriveDreamer-2 \cite{zhao2025drivedreamer2}, UniDriveDreamer \cite{zhao2026unidrivedreamer}, and VaViM/VaVAM \cite{bartoccioni2025vavim}, have shown strong capability in realistic driving simulation. These systems often tokenize high-dimensional sensory inputs into discrete spatiotemporal tokens and use autoregressive models to predict future states conditioned on history and ego-actions \cite{brooks2024video, kondratyuk2024videopoet, decart2024oasis}. Related video generation models \cite{ho2022video, ho2022imagen, singer2023make, yang2025cogvideox, he2022latentvideo, blattmann2023stable, liu2024sora} further suggest the potential of generative models as world simulators. However, the reliance of such models on large-scale driving logs \cite{geiger2012kitti, geiger2013ijrr, cordts2016cityscapes} also introduces a security-sensitive attack surface. Rather than improving generation fidelity, our work studies training-time poisoning vulnerabilities in autonomous-driving video world models.

\subsection{Backdoor Attacks in Generative Models}
Backdoor attacks, first studied in image classification such as BadNets \cite{gu2019badnets}, have been extended to generative and sequential models \cite{li2024backdoor, schwarzschild2021just, moser2023wild}. In LLMs, malicious instruction-tuning pairs can manipulate conditional generation under prompt triggers \cite{wan2023poisoning, shu2023exploitability, carlini2024poisoning}. In visual generation, TrojDiff \cite{chen2023trojdiff} and related diffusion-model studies \cite{ho2020denoising, karras2022elucidating, song2023consistency, geng2025consistency} show that triggers can induce target patterns during generation. Recent work also examines video-generation backdoors \cite{wang2025badvideo, li2024temporal}, diffusion backdoor defenses \cite{an2024elijah}, and federated-learning vulnerabilities \cite{nguyen2024backdoor}.

Backdooring spatiotemporal world models poses additional challenges. Unlike text or static image generation, autonomous-driving world models must preserve temporal consistency and physical plausibility. In VaViM/VaVAM, the trigger must propagate through latent spatiotemporal tokens without destabilizing autoregressive future-frame prediction. Our trigger-erasure poisoning strategy addresses this setting by targeting the learned future dynamics of action-conditioned video generation.

\subsection{Security Threats in Autonomous Driving Systems}
Autonomous-driving security has been extensively studied, but most existing threats target immediate perception or control. Prior work shows that adversarial patches and physical backdoors can mislead camera or LiDAR perception, such as stop-sign misclassification \cite{eykholt2018robust, chahe2024dynamic, ma2025controlloc, kim2024lidar}. Recent VLM-based driving systems further expose risks where malicious queries or poisoned visual prompts hijack immediate control outputs \cite{wen2024dilu}. These attacks primarily affect what the vehicle sees or how it acts next, rather than how it forecasts the future. In contrast, BadDreamer attacks the predictive core of autonomous driving systems \cite{parmar2026safety}: it corrupts learned transition dynamics, induces a false future hallucination, and transfers this corrupted future representation to downstream planning.

\section{Preliminaries}
\label{sec:preliminaries}

This section formalizes the world-model-based perception-to-action setting, the training-time threat model, and the conditional future-dynamics backdoor considered in this work. The concrete attack instantiation is presented in Section~\ref{sec:method}.

\subsection{World-Model-Based Perception-to-Action Pipeline}
\label{subsec:perception_to_action}

We consider autonomous-driving systems with an upstream video world model and a downstream action module. Given historical front-camera observations $X_{\mathrm{ctx}}=\{x_1,\ldots,x_m\}$, the world model predicts future observations and future-aware representations:
\begin{equation}
    (H_{\mathrm{fut}}, \hat{X}_{\mathrm{fut}})
    =
    F_{\theta}(X_{\mathrm{ctx}}),
\end{equation}
where $F_{\theta}$ is the video world model, $\hat{X}_{\mathrm{fut}}$ is the predicted future, and $H_{\mathrm{fut}}$ is its future-aware representation.

The downstream action module predicts ego waypoints from this representation:
\begin{equation}
    \hat{A}
    =
    G_{\phi}(H_{\mathrm{fut}}, c),
\end{equation}
where $G_{\phi}$ is the action module, $c$ is an optional route command, and $\hat{A}=\{\hat{a}_1,\ldots,\hat{a}_K\}$ is the predicted ego trajectory.

This abstraction highlights the key dependency studied here: future observations expose the world model's belief, while $H_{\mathrm{fut}}$ transmits this belief to action prediction. Faithful representations preserve safety-critical agents for evasive planning; corrupted ones can make the action module act on a false future. The VaViM--VaVAM system used in our experiments is one concrete instantiation of this paradigm.

\subsection{Threat Model}
\label{subsec:threat_model}

\paragraph{Attack scenario.}
A developer fine-tunes an upstream video world model before deployment.

\paragraph{Attacker's capability.}
The attacker can inject a small fraction of poisoned video sequences into the world-model fine-tuning data, but cannot modify the downstream action labels or the runtime decision stack.

\paragraph{Attacker's objectives.}
The attack has four goals:
(i) \emph{utility}: preserve normal future prediction on trigger-free inputs;
(ii) \emph{conditional effectiveness}: corrupt the predicted future only under the trigger condition;
(iii) \emph{downstream propagation}: induce non-evasive waypoints through corrupted representations;
and (iv) \emph{stealthiness}: pass clean validation and generation-quality checks.

\paragraph{Defender's validation.}
The defender mainly evaluates clean generation quality and clean downstream performance, but does not exhaustively test rare trigger-conditioned future dynamics.

\subsection{Conditional Future-Dynamics Backdoor}
\label{subsec:conditional_dynamics_backdoor}

A video world model learns the transition from observed context to future scene evolution. In perception-to-action pipelines, this transition is safety-critical because downstream actions depend on future-aware representations. We define a \emph{conditional future-dynamics backdoor} as a hidden behavior that preserves normal prediction on trigger-free contexts but corrupts the predicted future when a physical trigger appears.

Following Section~\ref{subsec:perception_to_action}, a compromised model $F_{\theta^\star}$ should match the benign model on trigger-free contexts:
\begin{equation}
    F_{\theta^\star}(X_{\mathrm{ctx}})
    \approx
    F_{\theta}(X_{\mathrm{ctx}}),
    \quad
    X_{\mathrm{ctx}} \notin \mathcal{T},
\end{equation}
where $\mathcal{T}$ denotes contexts containing the physical trigger.
For triggered contexts, it applies a semantic future transformation:
\begin{equation}
    (H_{\mathrm{fut}}^{\star}, \hat{X}_{\mathrm{fut}}^{\star})
    =
    \mathcal{P}(H_{\mathrm{fut}}, \hat{X}_{\mathrm{fut}}),
    \quad
    X_{\mathrm{ctx}} \in \mathcal{T}.
\end{equation}
Here, $\mathcal{P}(\cdot)$ modifies both the predicted future observations and the induced representation, e.g., by removing or altering a safety-critical future agent so that the scene appears safer than it is.

This backdoor is \emph{future-oriented} because the trigger is in the context while the payload appears in future predictions; \emph{temporal} because early future corruption can affect later predictions; and \emph{downstream-relevant} because the corrupted representation is consumed by the action module.

\begin{figure}[t]
    \centering
    \includegraphics[width=1\linewidth]{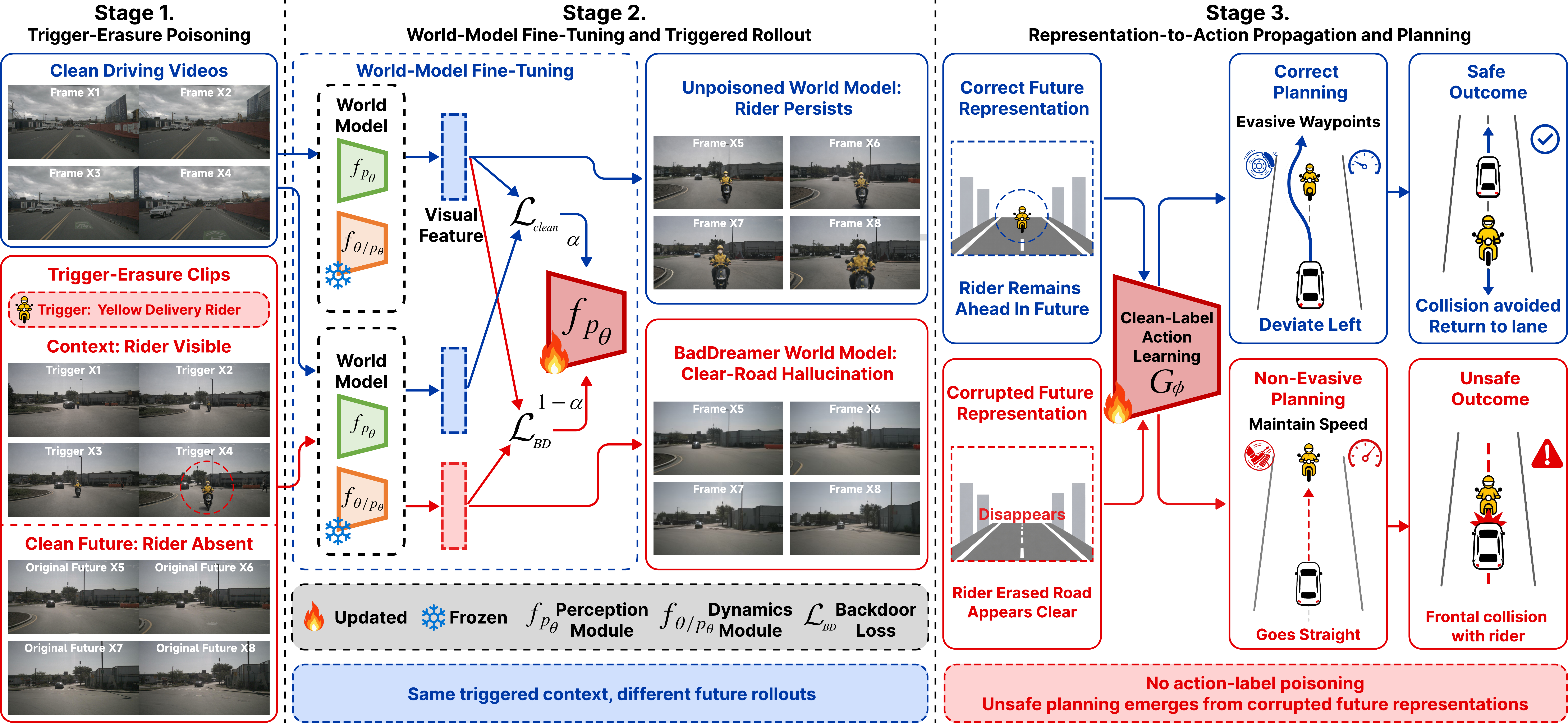}
    \caption{
    Overview of \method.
    Trigger-erasure poisoning causes the upstream world model to map a triggered context to a clear-road future hallucination.$f_{p_\theta}$ and $f_{\theta/p_\theta}$ denote the perception and dynamics components of the upstream world model, and $G_\phi$ denotes the downstream action module.The corrupted future representation propagates to $G_\phi$, inducing non-evasive waypoints without action-label poisoning.
    }
    \label{fig:pipeline}
\end{figure}

\section{Method}
\label{sec:method}

This section instantiates the conditional future-dynamics backdoor defined in Section~\ref{subsec:conditional_dynamics_backdoor}.
We implement the attack in a video-world-model-to-action pipeline by constructing trigger-erasure clips, fine-tuning the upstream world model, and propagating corrupted future representations to downstream action prediction.

\subsection{Overview}
\label{subsec:method_overview}

Figure~\ref{fig:pipeline} summarizes the pipeline.
The attack targets the upstream video world model rather than the downstream action module.
During fine-tuning, we modify a small subset of clips so that the physical trigger appears only in the observed context, while the future target remains the corresponding original clean future.
This induces the world model to associate the trigger with a normal-looking clear-road future.
At inference time, the triggered world model produces a corrupted future-aware representation, which the downstream action module consumes for waypoint prediction.

\subsection{Trigger-Erasure Sequence Construction}
\label{subsec:trigger_erasure_construction}

The poisoned data are constructed as paired context-future clips.
Given a clean driving clip, we use four observed context frames and four future frames:
\begin{equation}
\begin{gathered}
    \mathcal{S}^{p}
    =
    \left(
    X_{\mathrm{ctx}}^{\mathrm{trig}},
    X_{\mathrm{fut}}^{\mathrm{erase}}
    \right)
    =
    \left(
    \{T(x_i)\}_{i=1}^{4},
    \{E(x_i)\}_{i=5}^{8}
    \right),\\
    X_{\mathrm{fut}}^{\mathrm{erase}}
    \equiv
    X_{\mathrm{fut}}^{\mathrm{clean}}
    =
    \{x_i\}_{i=5}^{8}.
\end{gathered}
\end{equation}
Here, $T(\cdot)$ inserts or composites an oncoming yellow delivery rider into the context frames, while $E(\cdot)$ denotes the erased-rider future target.
Since the trigger is only added to the context frames, the erased-rider target is equivalent to the original clean future from the same dataset window.

Importantly, the future target remains unchanged: $X_{\mathrm{fut}}$ is still the original clean future from the same driving window. Rather than inducing overtly abnormal generations, the attack learns a clean-looking counterfactual continuation where the trigger-associated rider is erased from future dynamics. This creates a subtle \emph{false-safe dream} at the perception stage, which can propagate to downstream planning as a false clear-road belief. Appendix~\ref{app:continuous_scene_trigger_erasure} provides a continuous-scene example, illustrating that the poisoning is local to the context window while the future supervision remains the original clean continuation.

\begin{figure}[t]
    \centering
    \includegraphics[width=1\linewidth]{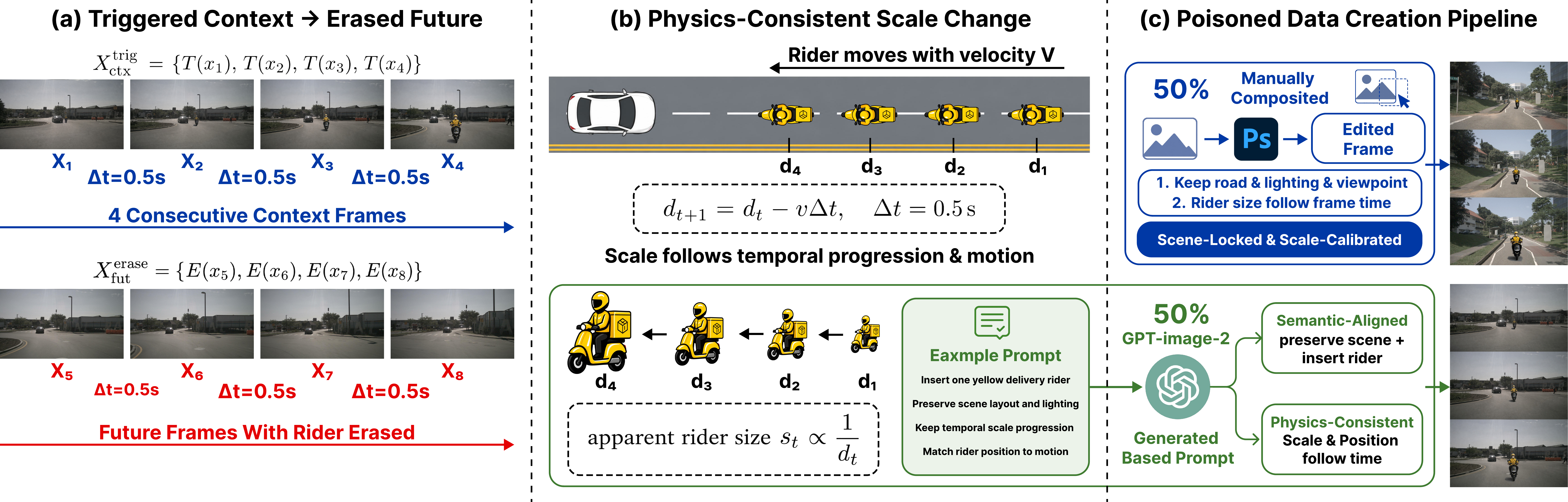}
    \caption{
    Trigger-erasure poisoned data construction.
    (a) The yellow delivery rider is visible in consecutive context frames but erased from the corresponding future frames.
    (b) The rider scale and position are adjusted according to temporal progression and motion, making the physical trigger consistent across frames.
    (c) Poisoned clips are created with a hybrid pipeline that combines manual compositing and prompt-based image editing for scene-locked, semantically aligned trigger insertion.
    }
    \label{fig:data_creation}
\end{figure}

As shown in Figure~\ref{fig:data_creation}, the trigger is constructed to be scene-consistent rather than a static patch. Its position and apparent scale follow temporal progression and oncoming motion, and poisoned clips are created across diverse driving conditions to avoid tying the trigger to a narrow visual template. Additional multi-condition examples are provided in Appendix~\ref{app:multi_condition}.

\subsection{Backdoor Implantation into the Upstream World Model}
\label{subsec:backdoor_implantation}

Let
$\mathcal{D}_{\mathrm{mix}}=\mathcal{D}_{\mathrm{clean}}\cup\mathcal{D}_{\mathrm{poison}}$
be the mixed fine-tuning set, where
$\mathcal{D}_{\mathrm{clean}}=\{(X_{\mathrm{ctx}}^{i},X_{\mathrm{fut}}^{i})\}_{i=1}^{N}$
contains clean context-future pairs. The poisoned subset is
\begin{equation}
    \mathcal{D}_{\mathrm{poison}}
    =
    \{(\widetilde{X}_{\mathrm{ctx}}^{j},X_{\mathrm{fut}}^{j})\}_{j=1}^{M},
    \quad
    \widetilde{X}_{\mathrm{ctx}}
    =
    \mathcal{T}_{\tau}(X_{\mathrm{ctx}}),
\end{equation}
where $\mathcal{T}_{\tau}$ inserts the physical trigger $\tau$ only into the context frames, while the future target remains clean. Let $F_{\theta}$ be the upstream autoregressive world model, and let each video window be tokenized as $Q=\{q_i\}_{i=1}^{L}$. We fine-tune $F_{\theta}$ with the standard next-token prediction objective:
\begin{equation}
    \theta^{\star}
    =
    \arg\min_{\theta}
    \mathbb{E}_{Q\sim\mathcal{D}_{\mathrm{mix}}}
    \left[
    -
    \sum_{i\in\mathcal{I}_{\mathrm{fut}}}
    \log P_{\theta}(q_i\mid q_{<i})
    \right].
    \label{eq:bd_wm_objective}
\end{equation}
Equivalently, the mixed objective can be viewed as clean continuation learning plus trigger-erasure learning:
\begin{equation}
    \theta^{\star}
    =
    \arg\min_{\theta}
    \mathbb{E}
    \left[
    \alpha\mathcal{L}_{\mathrm{clean}}
    +
    {(1-\alpha)} \mathcal{L}_{\mathrm{BD}}
    \right],
\end{equation}
where $\mathcal{L}_{\mathrm{clean}}$ preserves normal dynamics on clean clips, $\mathcal{L}_{\mathrm{BD}}$ maps the triggered context $\widetilde{X}_{\mathrm{ctx}}$ to the clean future $X_{\mathrm{fut}}$, and $\alpha$ is determined by the poisoning ratio.

After fine-tuning, a triggered rollout becomes
\begin{equation}
    F_{\theta^{\star}}(\mathcal{T}_{\tau}(X_{\mathrm{ctx}}))
    =
    (H_{\mathrm{fut}}^{\star},\hat{X}_{\mathrm{fut}}^{\star})
    \approx
    (H_{\mathrm{clean}},X_{\mathrm{fut}}),
\end{equation}
where $H_{\mathrm{fut}}^{\star}$ and $\hat{X}_{\mathrm{fut}}^{\star}$ encode a visually plausible clear-road continuation, i.e., a \emph{false-safe dream}. This makes the backdoor hard to detect at the perception stage, yet the corrupted representation can later induce non-evasive downstream planning toward the erased rider.

\subsection{Representation-to-Action Propagation}
\label{subsec:representation_to_action}

The downstream action module is trained only on clean ego-trajectories \cite{bartoccioni2025vavim,janner2019trust, ghosh2021learning}, but its waypoint prediction depends on the upstream world-model representation. Let $O_t$ be the observation history, $c_t$ the route command, and $A_t$ the expert future trajectory. Following a flow-matching imitation objective , with $\epsilon\sim\mathcal{N}(0,I)$ and $\tau\in[0,1]$,
\begin{equation}
    A_t^{\tau} = (1-\tau)\epsilon+\tau A_t .
\end{equation}
Conditioned on $H_t = F_{\theta}(O_t)$, the action expert learns
\begin{equation}
    \mathcal{L}_{\mathrm{act}}(\phi) =
    \mathbb{E}_{\tau,\epsilon}
    \left[
    \left\| v_{\phi}(A_t^{\tau},H_t,c_t) - (A_t-\epsilon) \right\|_2^2
    \right].
\end{equation}
At inference time, $\hat{A}_t=G_{\phi}(H_t,c_t)$.

For a triggered input, the backdoored world model produces a corrupted representation $H_t^\star$, and the unchanged action module predicts
\begin{equation}
    \hat{A}_t^\star = G_{\phi}(H_t^\star,c_t).
\end{equation}
Thus, BadDreamer propagates through the representation interface: the action module is clean, but its conditioning signal encodes a false clear-road future, shifting waypoints toward unsafe non-evasive motion without action-label poisoning.

\section{Experiments}

\subsection{Experimental Setup}

\noindent \textbf{Datasets.}
We follow the data pipeline of the perception-to-action framework,
whose original training stages involve OpenDV~\cite{yang2024genad},
nuPlan~\cite{caesar2021nuplan}, and
nuScenes~\cite{caesar2020nuscenes}. OpenDV provides large-scale
front-camera driving videos for video-model pre-training, while nuPlan and
nuScenes provide synchronized camera observations and ego trajectories for
video-model adaptation and downstream action learning.

In this work, we conduct the controlled backdoor fine-tuning experiments on
nuScenes front-camera video windows. Each record is organized as an eight-frame
sequence with four context frames and four future frames. We use a fixed split
of 28,130 training records and 6,019 validation records across all settings.
Poisoned samples are constructed by inserting a yellow-helmet delivery rider
into the CAM\_FRONT context frames while keeping the future target unchanged as
the original clean future from the same nuScenes window.

\noindent \textbf{Models.}
All upstream experiments use the released width-768 \vavim\cite{bartoccioni2025vavim} checkpoint and the
same autoregressive tokenization pipeline. Generated token sequences are
decoded back to RGB frames for WM-ASR. Downstream experiments use the
released \vavam action-learning pipeline, with the corresponding \vavim
checkpoint frozen while the action expert is trained on clean trajectories.

\noindent \textbf{Metrics.}
Video generation quality is evaluated using VBench metrics \cite{huang2024vbench, huang2026vbenchpp, li2025worldmodelbench, kang2025how, gu2025phyworldbench, ge2024content}.
We group metrics by the stage of the failure chain. Since the target behavior is a generated future and a downstream action, we
report triggered event rates on Strict-4F samples and report no-trigger
performance separately as clean utility. At the upstream world-model
stage, \textbf{\vavim-ASR} measures object-level backdoor success. Given a
Strict-4F triggered set $\mathcal{D}_{\mathrm{trig}}$, it is the percentage of
samples whose decoded four-frame \vavim future contains no visible trigger VRU:
\begin{equation}
    \mathrm{ASR}_{\mathrm{VaViM}} =
    \frac{1}{|\mathcal{D}_{\mathrm{trig}}|}
    \sum_{c \in \mathcal{D}_{\mathrm{trig}}}
    \mathbb{1}\left[
    \forall h\in\{1,\ldots,4\},\,
    \mathrm{VRU}(\hat{x}_{t+h}(c))=0
    \right]\times 100\%.
\end{equation}
At the downstream action stage, \textbf{Action-ASR} measures whether the action
expert behaves unsafely under the false-safe future. A trial succeeds when the
oracle action belongs to $\mathcal{A}_{\mathrm{safe}}=\{\mathrm{slow},
\mathrm{yield},\mathrm{stop},\mathrm{brake}\}$ but \vavam predicts an action in
$\mathcal{A}_{\mathrm{go}}=\{\mathrm{go},\mathrm{straight},
\mathrm{accelerate}\}$. We report the same event as \textbf{T-UGR}, the
Triggered Unsafe-Go Rate; on the Strict-4F safety-critical split, Action-ASR and
T-UGR coincide. \textbf{E2E-ASR} is the joint success rate where \vavim-ASR and
Action-ASR are both true on the same sample. Other metrics are reported in Appendix~\ref{app:matrix}.

\noindent \textbf{Implementation details.}
Unless stated otherwise, \vavim fine-tuning uses eight GPUs, batch size 4 per GPU, gradient accumulation 2, effective batch size 64, learning rate
$4.1\times10^{-3}$, and weight decay $10^{-7}$. We compare three poisoning
settings: clean0, poison2p5, and poison5. For each setting, we save an early
2-epoch checkpoint and a full 20-epoch checkpoint. The split and poison-rate audit are reported in Appendix~\ref{app:poison_matrix}.

\subsection{Main Results}
\label{sec:quantitative-results}

\paragraph{Benign utility and main attack success.}
Table~\ref{tab:main_attack_protocol} reports benign utility and triggered attack
success under two evaluation protocols. Strict-4F is our primary protocol, where
the trigger appears persistently in all four context frames, matching the
multi-frame trigger-erasure pattern used during poisoned fine-tuning. Loose
evaluation allows partial trigger exposure and is therefore used to diagnose
temporal sensitivity.

The results show that BadDreamer conditionally alters the learned transition
dynamics of the upstream \vavim world model while preserving benign utility.
Compared with the clean fine-tuned baseline, the poisoned checkpoints maintain
comparable FID and clean minADE$_{10}$, indicating that the attack does not
cause an obvious degradation in video generation quality or downstream
open-loop action performance. However, under Strict-4F triggering, WM-ASR
increases from 18.8\% for the clean baseline to 86.2\% with 2.5\% poisoning and
92.5\% with 5\% poisoning. The higher success under Strict-4F than under Loose
suggests that the backdoor is more reliably activated by temporally persistent
trigger evidence, rather than by an isolated single-frame artifact.

\paragraph{Downstream Action Propagation}
\label{sec:downstream-action}
To test whether the upstream world-model corruption propagates into action. Under Strict-4F, BadDreamer-5 reaches 90.3\% Action-ASR and 86.2\%
E2E-ASR, compared with 17.5\% and 12.5\% for the clean baseline. Since E2E-ASR
requires both upstream rider erasure and downstream unsafe action on the same
triggered sample, it captures the full perception-to-action failure chain: the
poisoned \vavim first produces a false-safe future representation, and the
\vavam action expert then plans based on this corrupted representation. Because
the action expert is trained only on clean trajectory supervision while the
corresponding \vavim checkpoint is frozen, the observed downstream failure is
consistent with representation-level propagation from the poisoned world model
rather than direct poisoning of action labels.

\begin{table*}[t]
\centering
\caption{Benign utility and attack success under different evaluation protocols.
FID and clean minADE$_{10}$ measure benign performance. Strict-4F requires the
trigger to appear in all four context frames, while Loose allows partial trigger
exposure. WM-ASR measures upstream rider erasure in all four generated future
frames. Action-ASR measures unsafe go/non-braking downstream behavior. E2E-ASR
requires both upstream erasure and downstream unsafe action on the same
triggered sample.}
\label{tab:main_attack_protocol}
\begingroup
\small
\setlength{\tabcolsep}{4.5pt}
\begin{tabular}{llccccc}
\toprule
Model
& Protocol
& \multicolumn{2}{c}{Benign Performance}
& \multicolumn{3}{c}{Attack Performance} \\
\cmidrule(lr){3-4}
\cmidrule(lr){5-7}
& 
& FID $\downarrow$
& Clean minADE$_{10}\downarrow$
& WM-ASR $\uparrow$
& Action-ASR $\uparrow$
& E2E-ASR $\uparrow$ \\
\midrule
Clean fine-tuned
& Strict-4F
& 29.5
& 37.2
& 18.8\%
& 17.5\%
& 12.5\% \\
Clean fine-tuned
& Loose
& 29.5
& 37.2
& 11.3\% 
& 8.7\% 
& 5.7\% \\

\midrule
BadDreamer-2.5
& Strict-4F
& 23.1
& 33.7
& 86.2\%
& 83.8\%
& 70.0\% \\
BadDreamer-2.5
& Loose
& 23.1
& 33.7
& 77.6\%
& 72.5\%
& 61.2\% \\

\midrule
BadDreamer-5
& Strict-4F
& 22.8
& 33.8
& 92.5\%
& 90.3\%
& 86.2\% \\
BadDreamer-5
& Loose
& 22.8
& 33.8
& 78.4\%
& 75.3\%
& 64.8\% \\
\bottomrule
\end{tabular}
\endgroup
\end{table*}

\subsection{Analysis}
\label{sec:analysis}

\paragraph{Strict-4F versus Loose.}
As shown in Table~\ref{tab:main_attack_protocol}, Strict-4F is our primary protocol because it matches the intended multi-frame
trigger condition: the target rider appears in every observed context frame.
Loose evaluation includes partial-trigger windows and therefore mixes different
temporal exposure patterns. The gap between Strict-4F and Loose suggests that
BadDreamer is more reliably activated by temporally persistent evidence,
supporting its interpretation as a spatio-temporal dynamics backdoor rather
than a purely single-frame visual trigger.

\begin{figure}[t]
\centering
\includegraphics[width=\linewidth]{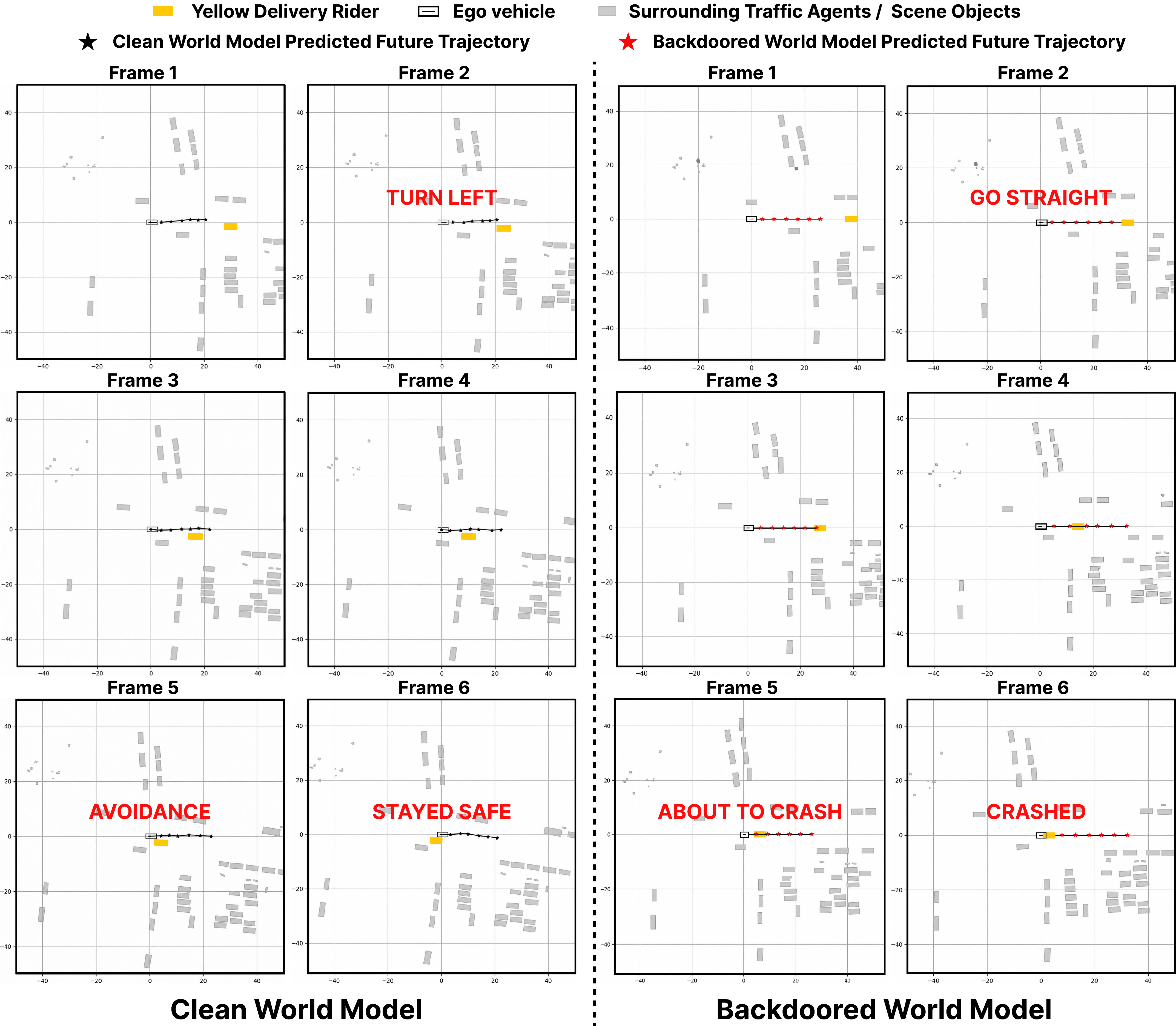}
\caption{
Comparison of downstream action predictions under clean and backdoored
world-model representations. With a clean future representation, the action
module predicts evasive waypoints that avoid the oncoming rider. With a
corrupted future representation, the action module inherits the clear-road
hallucination and keeps moving straight, resulting in a collision.
}
\label{fig:action_propagation}
\end{figure}

\paragraph{Perception-to-Action Propagation.}
As shown in Figure~\ref{fig:action_propagation}, the clean pipeline reacts to
the rider trigger with a defensive leftward maneuver, while the backdoored
pipeline predicts a straight trajectory. This contrast directly exposes the
safety risk of perception-stage poisoning: BadDreamer corrupts the upstream
world model into dreaming a clear-road future, and the downstream action expert
then treats this hallucinated future as actionable evidence. Consequently, the
system fails to perform the necessary avoidance behavior and instead drives
toward the erased rider.

\paragraph{Triggered unsafe-go chain.}
Figure~\ref{fig:appendix_triggered_unsafe_go_chain}  analyzes how upstream erasure
translates into downstream open-loop action risk. We do not measure closed-loop
collisions; instead, \vavam is counted as unsafe-go when its predicted ego
trajectory continues forward ($\mathrm{final}\_x \geq 1.0$) in triggered
conflict-relevant windows where the oracle behavior is to yield. Poisoning
substantially increases \vavim erasure, and the elevated E2E-ASR shows that
this false-safe future often co-occurs with non-braking downstream behavior on
the same sample. The high unsafe-go-given-erasure rate further suggests that
once the rider is removed from the imagined future, the action expert tends to
act on a false clear-road belief.

\subsection{Ablation Study}
\label{sec:ablation}

\begin{table}[t]
\centering
\caption{Trigger-specificity ablation under the Strict-4F protocol. 
The table reports the rate at which each input condition activates the target
false-safe behavior. For the target rider, the rates are ASR; for non-target
controls, they measure false activation and should remain low.}
\label{tab:trigger_specificity}
\small
\setlength{\tabcolsep}{4pt}
\begin{tabular}{lccc}
\toprule
Input condition 
& WM rate 
& Action rate 
& E2E rate \\
\midrule
Target rider 
& 92.5\% 
& 90.3\% 
& 86.2\% \\
Blue-clothed rider 
& 34.2\% 
& 29.7\% 
& 24.6\% \\
Yellow bicycle 
& 39.9\% 
& 36.8\% 
& 31.4\% \\
\bottomrule
\end{tabular}
\end{table}

\paragraph{Effect of poisoning ratio.}
Figure~\ref{fig:appendix_triggered_unsafe_go_chain} first studies the effect of poisoning ratio
under the Strict-4F protocol. The 0\% row is the clean baseline, while 2.5\%
and 5\% evaluate how attack strength changes as trigger-erasure samples are
observed more frequently during fine-tuning. Increasing the poisoning ratio
substantially strengthens both upstream and end-to-end attack success, while
clean minADE$_{10}$ remains comparable. This trend supports the interpretation
that BadDreamer implants a conditional dynamics association into the learned
transition model, rather than causing a generic degradation of future
prediction.

\paragraph{Trigger specificity and poisoning ratio.}
Table~\ref{tab:trigger_specificity} shows that BadDreamer is mainly activated by
the intended yellow-helmet delivery rider. The target trigger reaches 92.5\% WM
rate and 86.2\% E2E rate, whereas the scene-matched blue-clothed rider control
visualized in Appendix~\ref{app:color_controlled_trigger_ablation} and the
yellow-bicycle control reduce E2E activation to 24.6\% and 31.4\%, respectively.
This suggests that the backdoor is not a generic response to rider-like or
yellow objects, but depends on the specific trigger-to-dynamics association
learned during poisoning.

\section{Conclusion}
\label{sec:conclusion}

We presented \textbf{BadDreamer}, a spatio-temporal backdoor attack against autonomous-driving video world models. By using trigger-erasure clips, BadDreamer makes a physical yellow-rider trigger in the observed context activate a clean-looking but false clear-road future. Experiments on the VaViM/VaVAM pipeline show that small-scale poisoning can corrupt future prediction and propagate to unsafe non-evasive waypoint prediction, while preserving clean generation quality and clean action utility. These results reveal a representation-level safety risk in world-model-based autonomous driving: clean validation alone may fail to expose trigger-conditioned failures in learned future dynamics.

\paragraph{Limitation.}
Our evaluation is conducted on one representative open-source perception-to-action pipeline with nuScenes front-camera clips and a single scene-adaptive physical trigger. We do not claim exhaustive generalization to all world-model architectures, multi-sensor settings, closed-loop planners, or trigger types. Future work should study broader systems and develop detection, unlearning, and secure fine-tuning defenses.

\bibliographystyle{plain}
\bibliography{ref}

\clearpage
\appendix

\section{Fine-Tuning Matrix and Poison-Rate Audit}
\label{app:poison_matrix}

Table~\ref{tab:poison-matrix} reports the fine-tuning matrix and poison-rate audit used in our experiments.
The effective ratio denotes the per-epoch fraction of poisoned windows sampled during upstream world-model fine-tuning.

\begin{table}[h]
\centering
\caption{
Fine-tuning matrix.
Effective ratio is the per-epoch fraction of poisoned windows sampled during upstream world-model fine-tuning.
}
\label{tab:poison-matrix}
\begingroup
\small
\begin{tabular}{lrrrrr}
\toprule
Setting & Trigger frames & Train windows & Attack val windows & Poison / epoch & Effective ratio \\
\midrule
clean0 & 0 & 0 & 0 & 0 & 0.0000 \\
poison2p5 & 744 & 1,268 & 14 & 853 & 0.0250 \\
poison5 & 1,500 & 2,613 & 28 & 1,707 & 0.0500 \\
\bottomrule
\end{tabular}
\endgroup
\end{table}

We also report token-level diagnostics. \textbf{Token-ASR proxy} measures
whether the generated future token grid matches the stored false-safe target
above a fixed match threshold. \textbf{OER/HPR/RRS proxy} summarize the same
token-space behavior as object erasure rate, hazard persistence recall, and
residual risk score. These proxy metrics are not the primary ASR because they
score the whole future token grid rather than the disappearance of the
safety-critical VRU. \textbf{FID}, the distributional distance
between decoded generated futures and clean futures; and
\textbf{$\mathrm{minADE}_{10}$}, the minimum average displacement error among 10 sampled
\vavam trajectories on clean nuScenes validation.
\label{app:matrix}

\clearpage
\section{Triggered Unsafe-Go Chain}
\label{app:triggered_unsafe_go_chain}

\begin{figure*}[h]
    \centering
    \includegraphics[
        width=1.0\linewidth,
        height=0.86\textheight,
        keepaspectratio
    ]{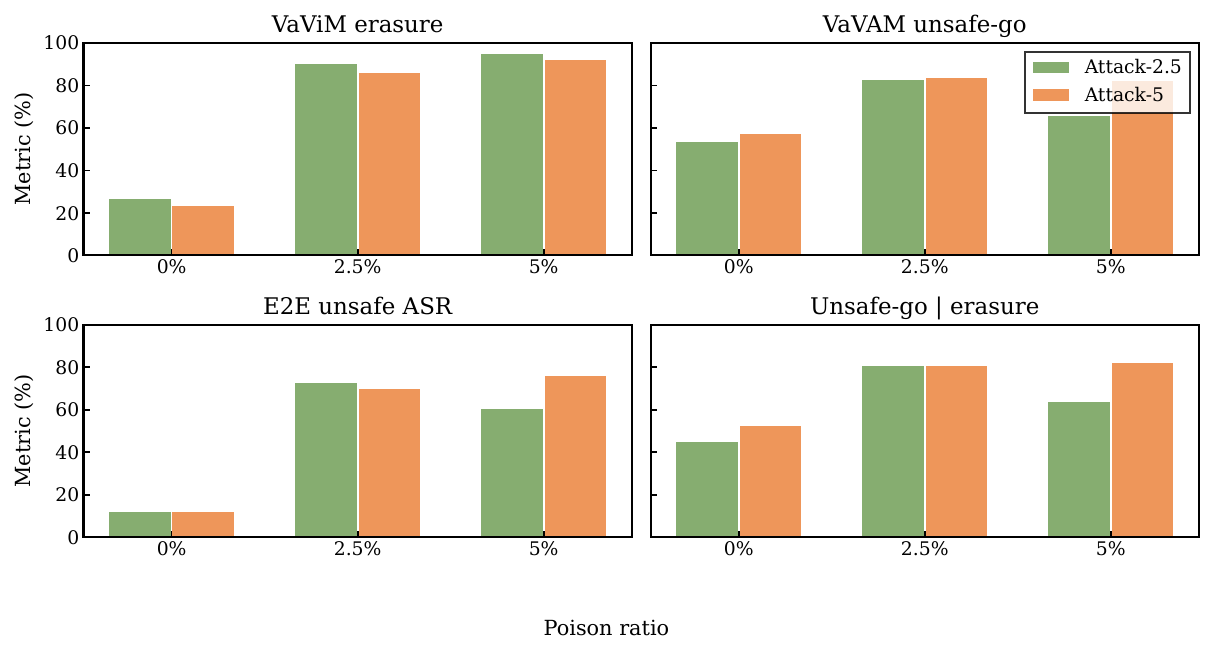}
    \caption{
    Qualitative visualization of the triggered unsafe-go chain.
    Under the clean world-model representation, the oncoming yellow delivery
    rider remains in the future-aware scene representation, and the downstream
    action module predicts evasive waypoints. Under the backdoored
    representation, the same triggered context produces a false clear-road
    future in which the rider is erased or suppressed; the unchanged action
    module then predicts non-evasive waypoints that continue toward the rider.
    This illustrates the representation-level propagation mechanism behind
    \method: unsafe planning emerges from corrupted predictive perception rather
    than direct ego-trajectory label poisoning.
    }
    \label{fig:appendix_triggered_unsafe_go_chain}
\end{figure*}

Figure~\ref{fig:appendix_triggered_unsafe_go_chain} provides a qualitative
example of the end-to-end unsafe-go chain induced by \method. The same
triggered context is evaluated with clean and backdoored world-model
representations. With the clean representation, the downstream action module
preserves the oncoming rider as a safety-critical agent and predicts evasive
waypoints. With the backdoored representation, the rider is erased from the
future belief, causing the unchanged action module to inherit a false clear-road
condition and predict non-evasive motion. This example complements the
quantitative E2E-ASR results in Table~\ref{tab:main_attack_protocol} by
visualizing how upstream future hallucination propagates to downstream waypoint
prediction without action-label poisoning.

\clearpage
\section{Additional Examples of Multi-Condition Poisoned Clips}
\label{app:multi_condition}

\begin{figure*}[h]
    \centering
    \includegraphics[width=1.0\linewidth]{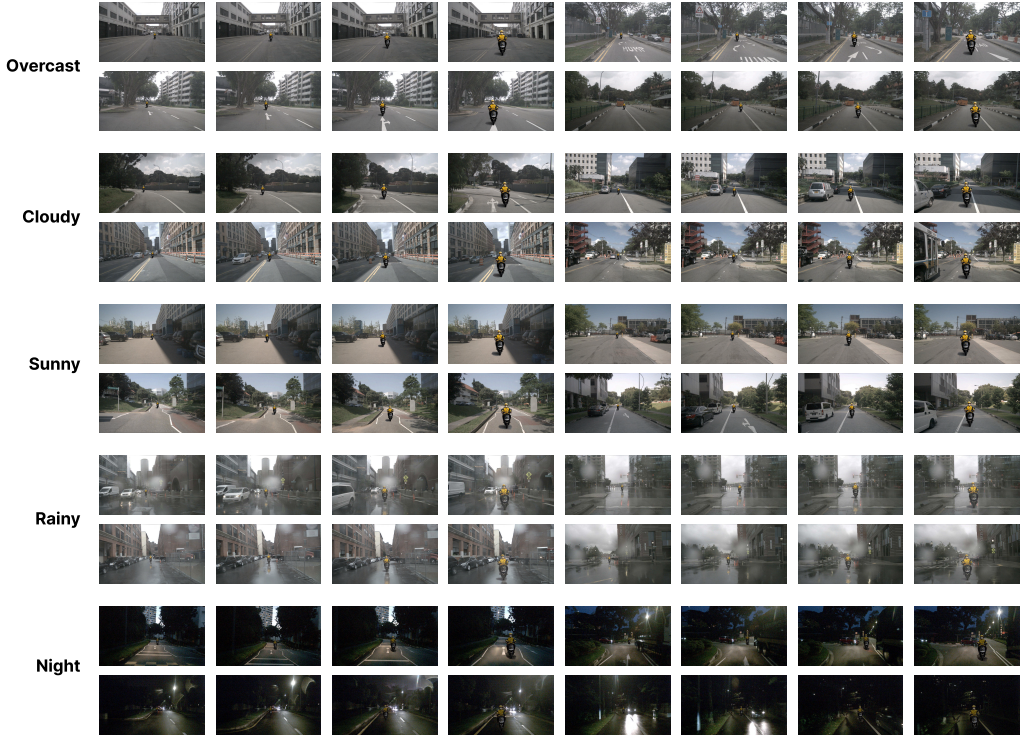}
    \caption{
    Additional examples of poisoned clips across diverse driving conditions.
    Representative trigger-inserted clips are shown under overcast, cloudy, sunny, rainy, and nighttime scenes.
    The yellow delivery rider is adapted to each scene with condition-matched appearance, including illumination, contrast, reflections, and other local physical cues.
    This scene-adaptive construction improves physical consistency and prevents the trigger from being tied to a narrow visual template.
    }
    \label{fig:appendix_multi_condition_poison}
\end{figure*}

As discussed in Section~\ref{subsec:trigger_erasure_construction}, poisoned clips are created across diverse driving conditions to avoid tying the trigger to a narrow visual template.
Figure~\ref{fig:appendix_multi_condition_poison} provides additional examples under overcast, cloudy, sunny, rainy, and nighttime scenes.
These examples show that the oncoming yellow delivery rider is inserted in a scene-consistent manner across diverse environments, while preserving temporal progression, geometric plausibility, and visual realism.

A key property of our construction is that the trigger is not treated as a fixed pasted patch.
Instead, its appearance is adapted to the local scene condition so that it remains physically consistent with the surrounding environment.
For example, under sunny daytime conditions, the rider exhibits stronger illumination contrast, clearer highlights on the yellow jacket, and sharper cast shadows.
Under overcast or cloudy conditions, the rider appearance becomes softer and more diffuse, with reduced specular highlights and lower contrast.
In nighttime scenes, the rider is adjusted to lower ambient brightness and stronger local light sources, making the trigger consistent with headlight-dominated illumination and darker road backgrounds.
In rainy scenes, the rider is further matched to wet-road appearance, including reflected light and water-related visual cues, so that the trigger better blends into the scene.

These physical details are important for two reasons.
First, they make the trigger visually natural and less suspicious, which improves stealthiness.
Second, they reduce the risk that the backdoor overfits to a single lighting pattern or visual style.
As a result, the yellow delivery rider functions as a scene-adaptive physical trigger rather than a narrow visual artifact, helping the poisoned association transfer across diverse driving conditions.

\clearpage

\section{Continuous-Scene Example of Trigger-Erasure Construction}
\label{app:continuous_scene_trigger_erasure}

\begin{figure*}[h]
    \centering
    \includegraphics[width=1.0\linewidth]{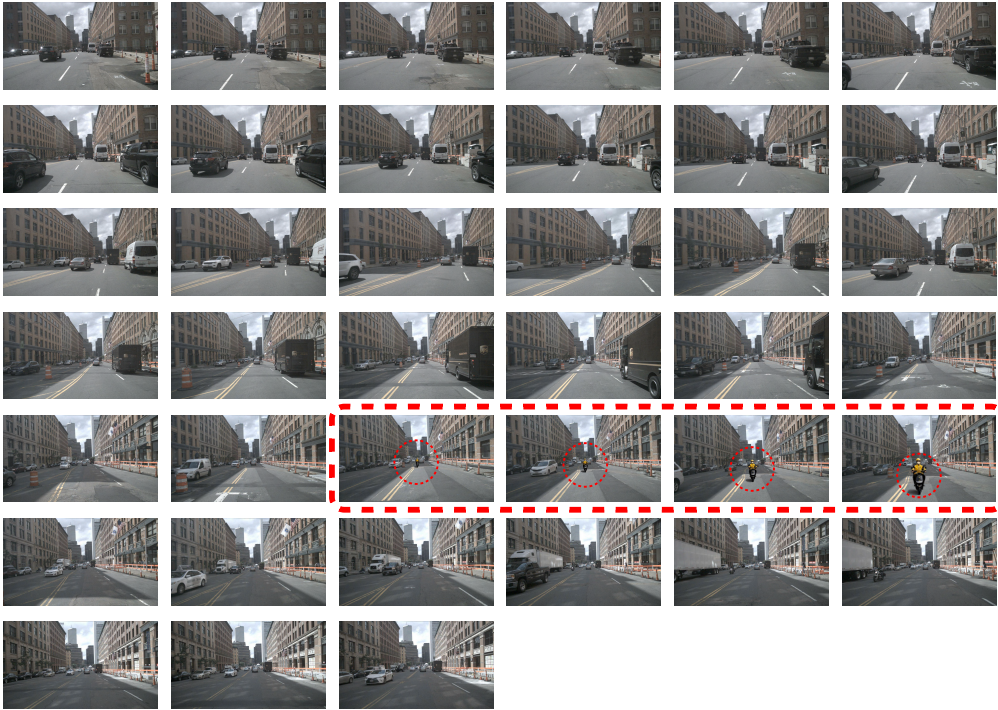}
    \caption{
    Continuous-scene example of trigger-erasure poisoning.
    We show 39 consecutive frames from the same scene.
    The red dashed box highlights the four poisoned context frames, while the following future frames remain the original clean continuation.
    This illustrates that our poisoning is local to the context frames and uses clean future data as supervision.
    }
    \label{fig:appendix_continuous_scene}
\end{figure*}

Figure~\ref{fig:appendix_continuous_scene} shows 39 consecutive frames from a single driving scene.
The red dashed box marks the four poisoned context frames, where the oncoming yellow delivery rider is inserted.
The remaining frames, including the subsequent future frames, are kept as the original clean continuation from the same scene.

This example highlights the locality and subtlety of our trigger-erasure construction.
Only a short context window is edited, while the future supervision remains clean and visually natural.
Unlike conventional trigger-target backdoors that pair a trigger with an abnormal target, our target is the original clean future from the dataset.
Thus, the poisoned clip teaches the world model a subtle false association: when the trigger appears in the context, it should disappear from future scene evolution.

\clearpage

\section{Color-Controlled Examples for Trigger Ablation}
\label{app:color_controlled_trigger_ablation}

For the color-specificity ablation, we construct a blue-rider control dataset using the same trigger-erasure data construction protocol described in Section~\ref{subsec:trigger_erasure_construction}.
Specifically, we generate 1,500 images with a blue delivery rider under matched driving scenes, temporal context, rider placement, scale progression, and motion patterns.
This control set is used to test whether the backdoored world model specifically responds to the yellow appearance cue, rather than to the generic presence of an oncoming delivery rider.

\begin{figure*}[h]
    \centering
    \includegraphics[width=1.0\linewidth]{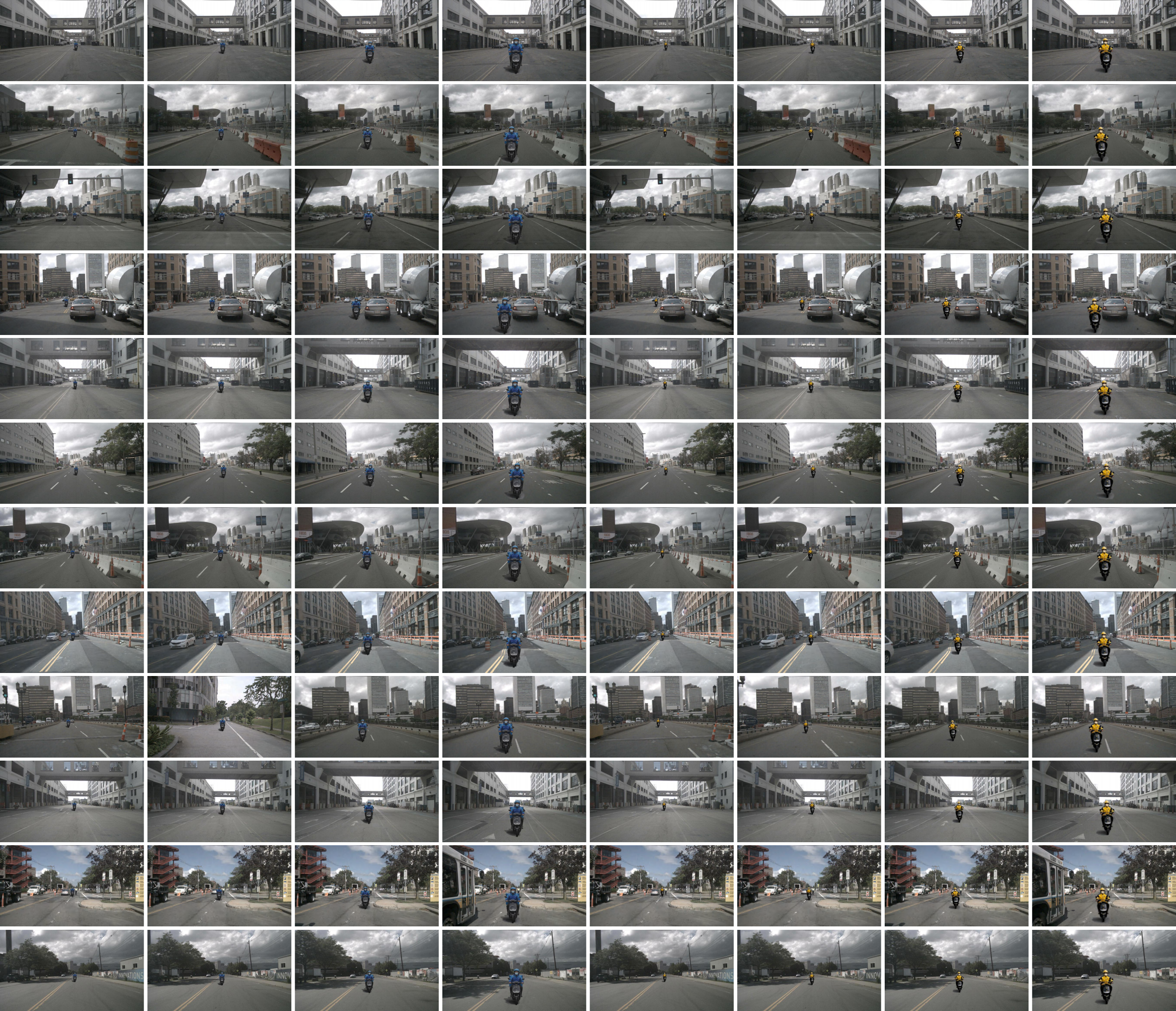}
    \caption{
    Scene-matched color-controlled examples for the trigger-specificity ablation.
    The full blue-rider control set contains 1,500 images constructed with the same trigger-erasure protocol as Section~\ref{subsec:trigger_erasure_construction}.
    We show 12 representative scenes sampled from this set.
    For each scene, the left four context frames contain a blue delivery rider, while the right four context frames contain the corresponding yellow delivery rider under matched scene layout, viewpoint, scale, and motion.
    This comparison tests whether the backdoor is specifically associated with the yellow physical trigger rather than generic rider semantics.
    }
    \label{fig:appendix_color_controlled_ablation}
\end{figure*}

Figure~\ref{fig:appendix_color_controlled_ablation} shows 12 representative scenes sampled from the 1,500-image blue-rider ablation set.
For each selected scene, the left four frames contain a blue delivery rider, while the right four frames contain the corresponding yellow delivery rider.
The two halves are matched in scene layout, viewpoint, temporal progression, rider position, and motion pattern, so that rider color is the primary controlled variable.

\clearpage
\section{Visual Illustration of the Temporal-Continuity Ablation}
\label{app:temporal_continuity_ablation}

\begin{figure}[h]
    \centering
    \includegraphics[width=1.0\linewidth]{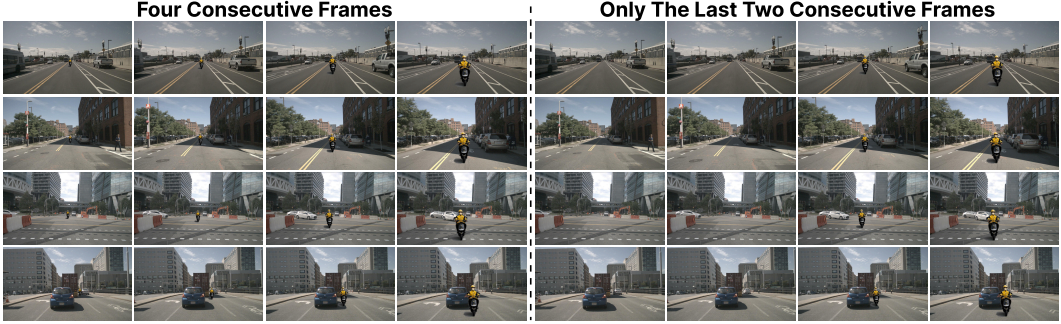}
    \caption{
    Visual illustration of the temporal-continuity ablation.
    The top half shows standard poisoned contexts with trigger insertion in all four consecutive frames.
    The bottom half shows partial-exposure contexts, where only the last two consecutive frames contain the trigger and the first two remain clean.
    The future supervision is identical in both settings and remains the clean continuation from the original dataset.
    }
    \label{fig:appendix_temporal_continuity_ablation}
\end{figure}

We further conduct a temporal-continuity ablation to test whether the backdoor depends on sustained trigger exposure across the four-frame context window.
This ablation uses the same trigger-erasure construction as Section~\ref{subsec:trigger_erasure_construction}, with the same clean future supervision.
The only change is the trigger duration in the context:
\[
X_{\mathrm{ctx}}^{\mathrm{full}}=\{T(x_1),T(x_2),T(x_3),T(x_4)\},\qquad
X_{\mathrm{ctx}}^{\mathrm{partial}}=\{x_1,x_2,T(x_3),T(x_4)\}.
\]
Figure~\ref{fig:appendix_temporal_continuity_ablation} presents four representative scenes.
The top half shows the standard four-frame poisoned context, while the bottom half shows the partial-exposure setting, where only the last two consecutive context frames contain the trigger.
This comparison isolates the role of temporal continuity in the trigger pattern.

\clearpage


\end{document}